\documentclass[runningheads]{llncs}

\usepackage{eccv}

\usepackage{eccvabbrv}

\usepackage{graphicx}
\usepackage{booktabs}
\usepackage{arydshln}
\usepackage[accsupp]{axessibility}  

\usepackage[table]{xcolor}

\def\redc{\bf\cellcolor[HTML]{FF999A}}
\def\orangec{\it \cellcolor[HTML]{FFCC99}}
\def\yellowc{\cellcolor[HTML]{FFF8AD}}

\usepackage{hyperref}

\usepackage{orcidlink}
\usepackage{multirow}

\newcommand{\safeincludegraphics}[2][]{%
    \IfFileExists{#2}{%
        \includegraphics[#1]{#2}%
    }{%
        \fbox{Missing figure: \texttt{\detokenize{#2}}}%
    }%
}

\begin{document}

\title{TIDES: Time-Derivative Event Simulation via Deformable Reconstruction} 

\titlerunning{TIDES}

\author{Christopher Thirgood, Dipon Kumar Ghosh, Simon Hadfield\\\
{\tt\small \{c.thirgood, d.ghosh, s.hadfield\}@surrey.ac.uk}}


\authorrunning{F.~Author et al.}

\institute{University of Surrey, Guildford, Surrey, UK}

\maketitle

\begin{figure}
    \centering
    \includegraphics[width=0.9\linewidth]{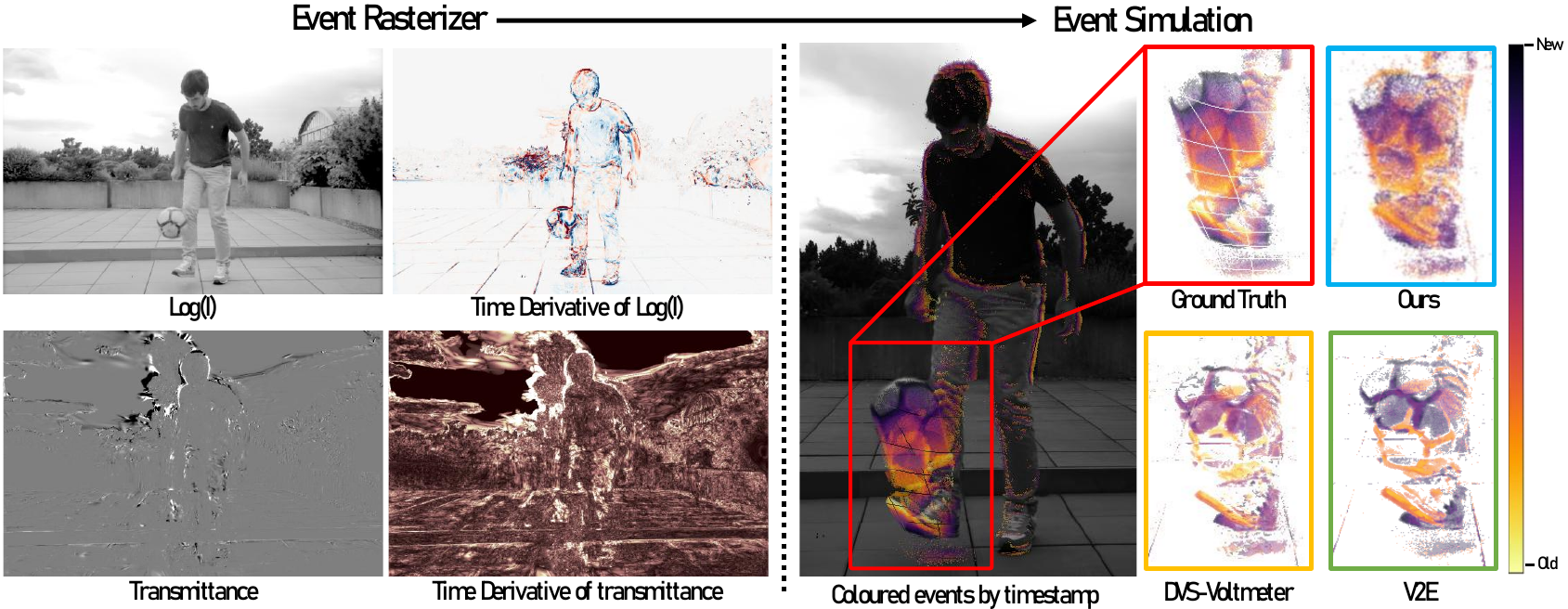}
    \vspace{-2mm}
    \caption{TIDES at a glance: a physically grounded, end-to-end event synthesis pipeline that couples consistent scene geometry with robustness to event bursting generation}
    \label{fig:teaserFig}
\end{figure}
\vspace{-4mm}

\begin{abstract}
Event cameras emit asynchronous events in response to environmental appearance changes.
The scarcity of real-world event datasets makes simulation essential.  
However, most simulators infer event timestamps from frame sequences, forcing many threshold crossings to share a small set of discrete times; a failure mode we term timestamp batching that worsens under fast motion and occlusion.

We present TIDES, a continuous-time event simulator built on dynamic Gaussian splatting.
Because TIDES operates on an explicit 3D scene representation with learnt geometry and motion, it can derive per-pixel intensity dynamics directly from the scene, rather than by differencing rendered frames. This enables accurate threshold-crossing prediction, including multiple crossings per rendering step, without temporal upsampling or frame interpolation.
The same 3D scene model reveals where objects partially occlude one another; TIDES uses this to guide adaptive time stepping, concentrating computation only in regions where occlusion dynamics make simple models of brightness change unreliable.

Finally, we model finite sensor bandwidth using a tile-level arbiter whose throughput, jitter, and event drops reproduce realistic sensor artifacts.
Across paired RGB-event benchmarks, TIDES attains state-of-the-art event-stream fidelity. We also show that events simulated by TIDES transfer more effectively to real downstream tasks than competitors'. 
\end{abstract}
\section{Introduction}
Event cameras measure visual change asynchronously by emitting events when the per-pixel log-intensity change exceeds a contrast threshold~\cite{Lichtsteiner2008DVS,Brandli2014DAVIS240,Gallego2022EventSurvey}. This sensing principle provides microsecond-scale temporal resolution and high dynamic range, enabling robust perception under fast motion and challenging illumination. These capabilities make it well suited for a wide range of computer vision applications \cite{GHOSH2024122743, GHOSH2025102891, KIM2023119917}.
Since collecting and labelling large event datasets is expensive, simulation is central for training and evaluation~\cite{Mueggler2017EventDatasetSimulator,Rebecq2018ESIM}.

Faithful event simulation remains difficult because event timestamps are a consequence of the 3D motion and visibility at each pixel, not of differences between discretely sampled frames. The dominant failure mode in event simulation is timestamp-batching, where many threshold crossings are forced to share a small set of timestamps determined by the rendering schedule. As illustrated in Fig.~\ref{fig:teaserFig}, the issue cannot be simply resolved by interpolating sub-frame crossing points based on brightness change. This is because the real 3D world is characterised by discrete object boundaries that move continuously through space, not by smooth global appearance changes.

The timestamp-batching issue has traditionally persisted even in fully synthetic pipelines, which theoretically have access to the exact 3D motion of the scene, because existing simulators use a rendered sequence of frames as their intermediate representation.
Rendering at extremely high rates can somewhat mitigate (but not fully remove) timestamp batching, but this becomes prohibitively slow and inaccurate in highly dynamic areas due to the reliance on temporal video interpolation models, which introduce artifacts.
%
%
Dynamic scenes amplify these issues. Finite-difference estimates of intensity slope mix distinct visibility regimes when occlusions or disocclusions occur between samples, yielding unstable local models precisely where accurate timing matters most~\cite{Rebecq2018ESIM,Han2024PECS}. 

In parallel, recent work in the field of Gaussian Splatting has provided mechanisms for real-time differentiable rendering with explicit visibility compositing. 4DGS extends these ideas to dynamic scenes via time-conditioned deformations~\cite{Kerbl2023GaussianSplatting,Wu2024FourDGS}. Because the scene is represented as an explicit collection of 3D primitives with known geometry and motion, the renderer provides direct access to which objects are visible, how they overlap, and how that visibility is changing. This is precisely the information that is most lacking in current frame-based event simulation methods.
However, a better 3D scene model alone is not enough. The simulator must translate scene-level geometry and motion into per-pixel rates of brightness changes that are consistent with the current visibility ordering, and must allocate computation adaptively to regions where objects overlap and simple brightness models break down.

To this end, we present TIDES, a continuous-time event simulator built on dynamic 4D Gaussian splatting~\cite{Wu2024FourDGS}. Rather than rendering frames and differencing them, TIDES differentiates the 3D-to-2D rendering process itself via Gaussian time dependent Jacobian vector products. Under TIDES custom Gaussian splatting rasterization method, these time-derivative rasterizations can be used to compute how fast each pixel's brightness is changing at any instant, while correctly accounting for which objects are visible. Concretely, a forward-mode rendering pass produces log-luminance $L$ and its instantaneous derivative $\dot{L}$ together with opacity dynamics $(\alpha,\dot{\alpha})$, all in the same compositing order as the primal image. This yields stable per-pixel threshold-crossing time prediction, including multiple crossings per step, without relying on a discretized rendering schedule.

Because the 3D scene model knows where objects overlap and how visibility is evolving, TIDES uses this information to guide adaptive time stepping, concentrating computation only where occlusion dynamics make a local linear model unreliable. Finally, TIDES models finite sensor bandwidth with a tile-level arbiter whose throughput and degradation are conditioned on the same scene-derived burst and visibility cues, reproducing burst-induced timestamp spreading and event drops. Overall, TIDES derives events from an explicit understanding of 3D scene structure and motion, rather than from frame differences.

We make the following contributions:
\begin{itemize}
    \item A forward-mode time-derivative renderer that produces visibility-consistent $(L,\dot{L})$, enabling continuous-time threshold-crossing prediction without temporal upsampling.
    \item Risk-guided adaptive stepping that concentrates computation where occlusion dynamics invalidate local linearity.
    \item An optional tile-level readout arbiter conditioned on renderer-derived burst cues, reproducing realistic timestamp spreading and drops.
    \item A motion-scaled spatiotemporal fidelity metric for event streams, evaluated alongside downstream transfer benchmarks.
\end{itemize}
\section{Related Work}

\subsection{Event simulators}
Most event simulators generate events by querying intensity at discrete times and estimating threshold-crossing timestamps by interpolating changes in $\log I$. ESIM popularized this renderer-coupled design and remains a common baseline due to its simplicity and scalability \cite{Rebecq2018ESIM}. However, timestamp structure is fundamentally constrained by the sampling schedule, so fast motion and visibility changes can collapse many events onto a small set of timestamps even when the comparator model is otherwise correct.

A complementary line of work improves simulator realism by modeling sensor stages beyond an ideal comparator. v2e composes photoreceptor-style dynamics with leak, threshold mismatch, background activity, and timestamp noise to better match device behavior \cite{Hu2021v2e}. V2CE~\cite{Zhang2024V2CE} extends this with a learned noise model calibrated to specific hardware. Circuit-inspired models such as DVS-Voltmeter provide a more mechanistic account of event generation and show how stochastic dynamics affect temporal statistics \cite{Lin2022DVSVoltmeter}. These approaches improve noise and readout realism, but when driven by discretely sampled video or renders, timestamp batching persists because event times are still inferred from under-sampled intensity signals.

Even with realistic pixel dynamics, temporal fidelity can fail if readout constraints are ignored. ICNS and related characterization work show that shared readout and arbitration can impose structured delays and drops under bursts, motivating explicit bottleneck models rather than treating timestamps as ideal crossing times \cite{Joubert2021CharacterizedSim}. Such models, however, still rely on the upstream simulator to generate realistic continuous-time bursts as input to the arbiter.

Physically grounded analytical pipelines such as PECS aim to close the realism gap by modeling image formation and sensor behavior more explicitly, including lens simulation and multispectral rendering \cite{Han2024PECS}. The trade-off is practical complexity and compute\cite{thirgood2025hydra, thirgood2026featureslam, thirgood2023raspectloc}, since accurate timing can require dense temporal querying and physics-heavy rendering is difficult to scale or integrate into modern learning workflows.

Our work targets the common bottleneck across these analytical simulator families: producing faithful continuous-time timestamps under non-linear visibility, without brute-force high-rate rendering. We introduce an event rasterization module that renders instantaneous temporal derivatives and visibility-consistent cues to directly support continuous-time threshold-crossing prediction, and can optionally be paired with explicit arbitration models.

\subsection{Dynamic neural rendering as a substrate for event simulation}
Neural rendering offers a convenient intermediate representation for event data because it supports view-dependent image formation and can be queried at arbitrary poses and times. Most prior event-based NeRF work addresses the inverse problem, learning radiance fields from events for reconstruction and novel-view synthesis~\cite{Rudnev_2023_CVPR,Hwang_2023_WACV,Klenk2023ENeRF,Low_2023_ICCV,Feng2025AENeRF}, with extensions to dynamic scenes via time-conditioned or deformation-based fields~\cite{Bhattacharya_2024_WACV,Rudnev_2025_CVPR}. While relevant as background, volumetric NeRF rendering is typically too slow for event-rate simulation and exposes limited visibility diagnostics near occlusions.

Gaussian splatting provides real-time differentiable rendering with explicit visibility compositing~\cite{Kerbl2023GaussianSplatting, thirgood2025hypergs}, and 4DGS extends it to dynamic scenes via time-conditioned deformations~\cite{Wu2024FourDGS}. Recent event-based splatting work similarly focuses on the inverse direction, recovering Gaussian scene representations from events (sometimes with auxiliary frames) for reconstruction and novel-view rendering~\cite{Han2024Event3DGS,DBLP:conf/corl/XiongWHFAHM24,Yura_2025_CVPR,Zahid2025E3DGS,Feng2025E4DGS}. In contrast, we are the first to study the forward problem; using a learned 4DGS scene as input, and simulating events by predicting continuous-time threshold crossings from renderer-native dynamics and visibility signals.
\section{Method}
\label{sec:method}

\begin{figure}
    \centering
    \includegraphics[width=\linewidth]{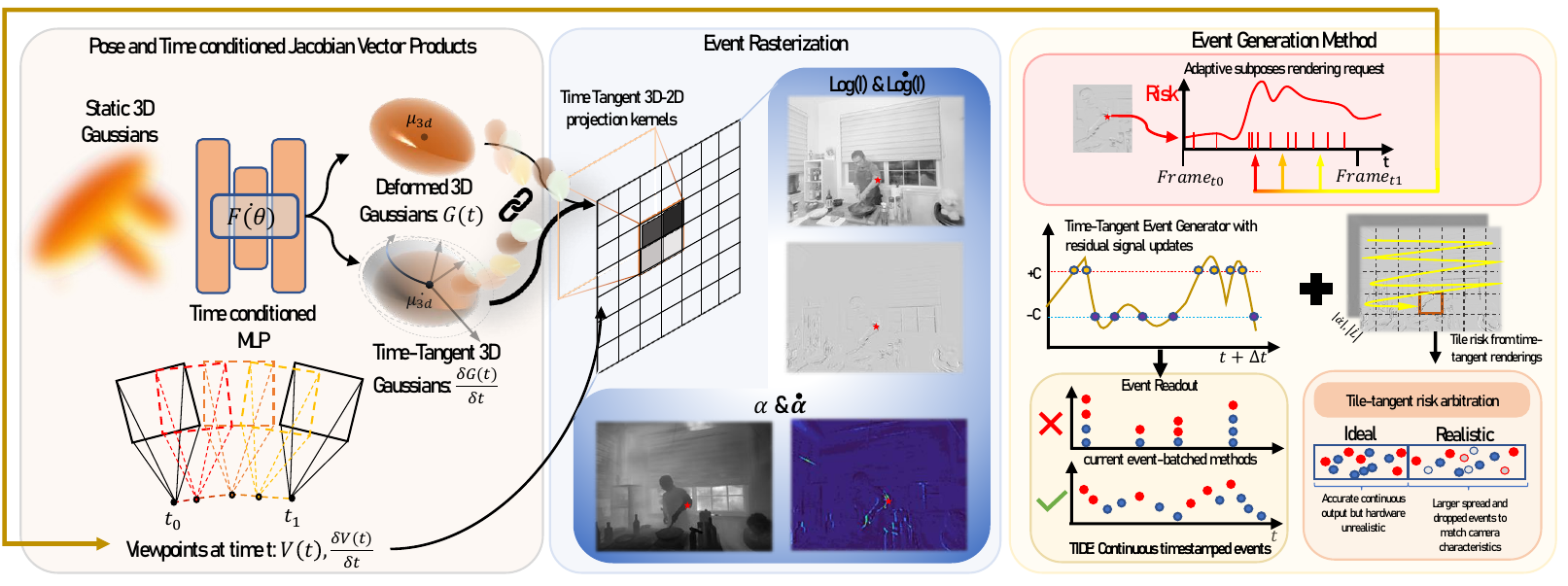}
    \caption{System diagram for TIDES. A primal and time-derivative splatting pass outputs $(L,\dot{L},\alpha,\dot{\alpha})$ with visibility-consistent compositing. These dynamics drive continuous-time threshold-crossing event times, avoid timestamp batching, and trigger risk-guided adaptive sub-posing under high motion and mixed visibility. An optional tile-level arbiter conditions readout spreading and drops on the same burst and visibility cues.}
    \label{fig:system_diagram}
    \vspace{-1em}
\end{figure}

The central design goal of TIDES is to predict event timestamps from an understanding of the underlying 3D structure and motion of the environment, rather than from finite differences between discretely rendered frames. To this end, our event rasterizer takes a continuous-time dynamic Gaussian splatting scene (fit from RGB videos), and produces an event stream by solving per-pixel contrast threshold crossings in continuous time. Between consecutive camera timestamps, we advance time with an adaptive step $\Delta t$ and query a time-derivative splatting renderer. At each query time $t$ and pixel location $\mathbf{u}$ TIDES outputs log-luminance $L(\mathbf{u},t)$ and its instantaneous derivative $\dot{L}(\mathbf{u},t)$. It also outputs opacity $\alpha(\mathbf{u},t)$ and its derivative $\dot{\alpha}(\mathbf{u},t)$, as well as lightweight accumulation diagnostics $(b(\mathbf{u},t),c(\mathbf{u},t))$. Crucially, both $\dot{L}$ and $\dot{\alpha}$ are differentiated under the active front-to-back compositing order, yielding visibility-consistent instantaneous dynamics. This makes the threshold-crossing timestamp estimation a well-posed root-finding problem and provides renderer-native signals for adaptive stepping and readout modelling.

\subsection{Dynamic Gaussian splatting image formation}
\label{sec:gs_rendering}

We use EWA Gaussian splatting and front-to-back compositing as in 3DGS~\cite{Kerbl2023GaussianSplatting}, with time-dependent parameters as in dynamic Gaussian representations~\cite{Wu2024FourDGS}. At time $t$, each Gaussian $k$ induces a screen-space elliptical kernel with projected mean $\mathbf{u}_k(t)$ and precision (inverse covariance) $\mathbf{Q}_k(t)$. For rendered pixel $\mathbf{u}$, the unnormalized contribution from Gaussian $k$ is
\begin{equation}
g_k(\mathbf{u},t)=\exp\!\Big(-\tfrac12(\mathbf{u}-\mathbf{u}_k(t))^\top\mathbf{Q}_k(t)(\mathbf{u}-\mathbf{u}_k(t))\Big).
\label{eq:gauss_weight}
\end{equation}
We define the per-splat opacity contribution
\begin{equation}
\tau_k(\mathbf{u},t)=\mathrm{clamp}\big(\sigma_k(t)\,g_k(\mathbf{u},t),\,0,\,\tau_{\max}\big),
\label{eq:tau_def}
\end{equation}
where $\sigma_k(t)$ is the Gaussian's time dependent opacity and $\tau_{\max}<1$ is a fixed saturation constant for numerical stability.
To find the Gaussians' final visibility weights $w_k$, their transmittance is composited in front-to-back order, with transmittance $T_0(\mathbf{u},t)=1$,
\begin{equation}
w_k(\mathbf{u},t)=T_{k-1}(\mathbf{u},t)\,\tau_k(\mathbf{u},t),
\qquad
T_k(\mathbf{u},t)=T_{k-1}(\mathbf{u},t)\big(1-\tau_k(\mathbf{u},t)\big).
\label{eq:ewa_weights}
\end{equation}
Thus, the rendered RGB ($\mathbf{I}$) and opacity ($\alpha$) are combined across Gaussians as
\begin{equation}
\mathbf{I}(\mathbf{u},t)=\sum_k w_k(\mathbf{u},t)\,\mathbf{c}_k(t)+T_K(\mathbf{u},t)\,\mathbf{I}_{bg}(t),
\qquad
\alpha(\mathbf{u},t)=1-T_K(\mathbf{u},t).
\label{eq:composite_rgb}
\end{equation}
Where $\mathbf{c}_k(t)$ is the time-dependent color of Gaussian $k$ and $\mathbf{I}_{bg}$ is the time-dependent background color.
We convert to luminance $Y(\mathbf{u},t)=\mathbf{w}^\top\mathbf{I}(\mathbf{u},t)$ using channel-weightings $\mathbf{w}$ and define the log signal that defines event-generation
\begin{equation}
L(\mathbf{u},t)=\log\!\big(Y(\mathbf{u},t)+\epsilon\big).
\label{eq:logI}
\end{equation}
This compositing chain defines the visibility ordering; TIDES evaluates time dynamics under the same ordering to obtain visibility-consistent event timings.
We provide the full projection Jacobians and screen-space covariance expressions used to compute $\mathbf{u}_k(t)$ and $\mathbf{Q}_k(t)$ in the supplementary material.
\subsection{Forward-mode time-derivative Gaussian splatting}
\label{sec:jvp_derivatives}

Accurate crossing-time prediction requires an instantaneous log-luminance slope that is consistent with renderer visibility. 
To be clear, the volumetric rendering process described above is differentiable. However, the derivatives of the rendering process itself are not what drives event-generation. Rather, we must first extract the derivatives of the scene parameters with respect to time, and then propagate them through the derivatives of our rendering process. This allows us to understand analytically and explicitly, how the final rendered appearance will change with respect to time ($\dot{L}$), instead of approximating it by differencing two rendering outputs.


More formally, a learned deformation model maps time to Gaussian parameters
\begin{equation}
\Theta(t)=\big(\boldsymbol{\mu}(t),\mathbf{s}_{\mathrm{raw}}(t),\mathbf{q}_{\mathrm{raw}}(t),\sigma_{\mathrm{raw}}(t),\mathbf{a}(t)\big),
\qquad \Theta:\mathbb{R}\rightarrow\mathbb{R}^{P\times d},
\end{equation}
where $P$ is the number of Gaussians.
This model is differentiable, so we recover the time derivatives $\dot{\Theta}(t) \triangleq \frac{d}{dt}\Theta(t)$ via a Jacobian-vector product (JVP) along the time axis, yielding per-Gaussian $(\dot{\boldsymbol{\mu}}_k,\dot{\mathbf{s}}_{\mathrm{raw},k},\dot{\mathbf{q}}_{\mathrm{raw},k},\dot{\sigma}_{\mathrm{raw},k},\dot{\mathbf{a}}_k)$.
This gives exact time derivatives in one evaluation, avoiding the inaccuracy of finite difference approximation, while remaining lightweight compared to reverse-mode differentiation through the rasterizer.

We compute this Jacobian–vector product in forward-mode automatic differentiation through the deformation network, not by finite differences in time.
Because the renderer applies element-wise activations to the raw deformation outputs, we propagate the time derivatives through these activations analytically via the chain rule, for example
\begin{align}
\mathbf{s}_k &= \exp(\mathbf{s}_{\mathrm{raw},k}), & \dot{\mathbf{s}}_k &= \mathbf{s}_k \odot \dot{\mathbf{s}}_{\mathrm{raw},k}, \\
\sigma_k &= \mathrm{sigmoid}(\sigma_{\mathrm{raw},k}), & \dot{\sigma}_k &= \sigma_k(1-\sigma_k)\,\dot{\sigma}_{\mathrm{raw},k}.
\end{align}
Quaternions are normalized before rasterization, and we propagate $\dot{\mathbf{q}}_k$ consistently through this normalization.

If the camera is nonstatic, we treat the camera parameters as time-dependent renderer inputs. When ground truth camera time derivatives $\dot{\mathbf{V}}(t),\dot{\mathbf{P}}(t),\dot{\mathbf{c}}(t)$ are not available, they are estimated from neighboring camera timestamps. Note that this does not reintroduce finite differences for contrast slopes, since $\dot{L}$ is still computed analytically by differentiating the resulting renderer traversal itself.

The time-tangent renderer computes per-Gaussian screen-space primitives $(\mathbf{u}_k,\dot{\mathbf{u}}_k,\mathbf{Q}_k,\dot{\mathbf{Q}}_k)$ and appearance $(\mathbf{c}_k,\dot{\mathbf{c}}_k)$, then composites them in the same front-to-back order as the primal pass. This ensures that temporal derivatives are consistent with visibility decisions such as early termination, clamping, and occlusion ordering.
In particular, $\dot{\mathbf{c}}_k$ includes derivatives through the view direction $\mathbf{v}_k(t)$ and the spherical-harmonic basis, and therefore captures motion-induced view-dependent appearance changes.

We now derive $\dot{L}$ by differentiating each stage of the compositing chain in Sec.~\ref{sec:gs_rendering}, starting from per-Gaussian screen-space contributions and propagating through front-to-back blending.
Using Eq.~\ref{eq:gauss_weight}, let
\begin{equation}
p_k(\mathbf{u},t)=-\tfrac12\,\mathbf{d}_k^\top\mathbf{Q}_k\,\mathbf{d}_k,
\qquad
\mathbf{d}_k=\mathbf{u}-\mathbf{u}_k(t).
\end{equation}
Since $\mathbf{u}$ is fixed, $\dot{\mathbf{d}}_k=-\dot{\mathbf{u}}_k$ and
\begin{equation}
\dot{p}_k
=
-\tfrac12\Big(
\mathbf{d}_k^\top\dot{\mathbf{Q}}_k\,\mathbf{d}_k
+2\,\dot{\mathbf{d}}_k^\top\mathbf{Q}_k\,\mathbf{d}_k
\Big),
\qquad
\dot{g}_k=g_k\,\dot{p}_k.
\label{eq:power_dot}
\end{equation}
Using Eq.~\ref{eq:tau_def}, define $\tau_k^{\mathrm{raw}}=\sigma_k g_k$ and
\begin{equation}
\dot{\tau}^{\mathrm{raw}}_k=\dot{\sigma}_k\,g_k+\sigma_k\,\dot{g}_k,
\qquad
\dot{\tau}_k=
\begin{cases}
0 & \text{if } \tau_k \text{ is saturated at } \tau_{\max},\\
\dot{\tau}^{\mathrm{raw}}_k & \text{otherwise.}
\end{cases}
\label{eq:tau_dot}
\end{equation}
With per-Gaussian derivatives in hand, we differentiate the front-to-back compositing.
Differentiating Eq.~\ref{eq:ewa_weights} yields
\begin{align}
w_k &= T_{k-1}\tau_k, &
\dot{w}_k &= \dot{T}_{k-1}\tau_k + T_{k-1}\dot{\tau}_k,
\label{eq:w_dot_clean}\\
T_k &= T_{k-1}(1-\tau_k), &
\dot{T}_k &= \dot{T}_{k-1}(1-\tau_k) - T_{k-1}\dot{\tau}_k.
\label{eq:T_dot_clean}
\end{align}
The final opacity derivative follows from $\alpha=1-T_K$ as $\dot{\alpha}=-\dot{T}_K$.
%
Finally, differentiating Eq.~\ref{eq:composite_rgb} yields the time derivative of the rendered image
\begin{equation}
\dot{\mathbf{I}}
=
\sum_k\big(\dot{w}_k\,\mathbf{c}_k+w_k\,\dot{\mathbf{c}}_k\big)
+\dot{T}_K\,\mathbf{I}_{bg}+T_K\,\dot{\mathbf{I}}_{bg}.
\label{eq:rgb_dot_clean}
\end{equation}
With luminance $Y=\mathbf{w}^\top\mathbf{I}$ and $\dot{Y}=\mathbf{w}^\top\dot{\mathbf{I}}$, we obtain
\begin{equation}
\dot{L}(\mathbf{u},t)=\frac{\dot{Y}(\mathbf{u},t)}{Y(\mathbf{u},t)+\epsilon}.
\label{eq:logI_dot}
\end{equation}

\subsection{Comparator and sensor model}
\label{sec:event_generation}
We next extend $(L,\dot{L})$ from the time-derivative renderer to define a per-pixel driving signal $S(\mathbf{u},t)=\mathcal{F}(L(\mathbf{u},t))$, where $\mathcal{F}$ is an optional sensor-front-end operator that models pixel-level analog processing. By default $\mathcal{F}$ is identity, so $S=L$ and $\dot{S}=\dot{L}$ directly. When enabled, $\mathcal{F}$ composes standard sensor-modelling blocks: optical Point-Spread-Function blur (applied to both $L$ and $\dot{L}$), a first-order photoreceptor low-pass, 
a center-surround spatial filter, and additive leak or noise terms. This sensor front-end is orthogonal to our continuous-time timing formulation.

We adopt a residual-state comparator. Each pixel maintains a reference $S_{\mathrm{ref}}(\mathbf{u})$, initialized to $S(\mathbf{u},t_0)$ and updated at each firing. An event is triggered when the signal departure from this reference reaches a contrast threshold: $S(\mathbf{u},t)-S_{\mathrm{ref}}(\mathbf{u})=+C^{+}_{\mathbf{u}}$ for a positive event, or $=-C^{-}_{\mathbf{u}}$ for a negative one. 
In the simplest case both polarities share a single global threshold $C^{+}_{\mathbf{u}}=C^{-}_{\mathbf{u}}=C$. To model realistic sensor variation we optionally allow asymmetric thresholds ($C^{+}\neq C^{-}$) and per-pixel jitter by drawing $C^{+}_{\mathbf{u}},C^{-}_{\mathbf{u}}$ independently around nominal values; once sampled, thresholds remain fixed for the simulation. We optionally also model comparator leak as a slow reference drift $S_{\mathrm{ref}} \leftarrow S_{\mathrm{ref}} - \lambda_{\mathrm{leak}}\,\Delta t$.

Within a simulation window $(t_0,t_0+\Delta t]$ we approximate $S$ locally. The first-order (linear) model is
\begin{equation}
S(\mathbf{u},t_0+\delta)\approx S_0(\mathbf{u})+\dot S_0(\mathbf{u})\,\delta,
\qquad \delta\in(0,\Delta t],
\label{eq:linear_local_model}
\end{equation}
where $S_0=S(\cdot,t_0)$ and $\dot S_0=\dot S(\cdot,t_0)$. Because $\dot S_0$ is obtained by analytic differentiation of the rendering pipeline, it inherits per-pixel visibility ordering and reflects occlusion and disocclusion dynamics without finite-difference approximation. When a second render $S_1=S(\cdot,t_0+\Delta t)$ is available, we optionally fit a quadratic local model using $S_0$, $\dot S_0$, and $S_1$, improving timing under signal curvature.

Given the local model, we solve for crossing times. Under the linear model with $\dot S_0(\mathbf{u})\neq 0$, the candidates are
\begin{equation}
\delta_{+}(\mathbf{u})=\frac{S_{\mathrm{ref}}(\mathbf{u})+C^{+}_{\mathbf{u}}-S_0(\mathbf{u})}{\dot S_0(\mathbf{u})},
\quad
\delta_{-}(\mathbf{u})=\frac{S_{\mathrm{ref}}(\mathbf{u})-C^{-}_{\mathbf{u}}-S_0(\mathbf{u})}{\dot S_0(\mathbf{u})}.
\label{eq:delta_candidates}
\end{equation}
Under the quadratic model, the candidates are the smallest positive roots of the corresponding quadratic in $(0,\Delta t]$. We consider $\delta_{+}$ only when $\dot S_0>0$ and $\delta_{-}$ only when $\dot S_0<0$, and accept a candidate only if $\delta\in(0,\Delta t]$. The earliest valid crossing $\delta^{*}$ yields the event timestamp $t^{*}=t_0+\delta^{*}$. We append $(t^{*},\mathbf{u},+1)$ to the output stream $\mathcal{E}$ and set $S_{\mathrm{ref}}\leftarrow S_{\mathrm{ref}}+C^{+}_{\mathbf{u}}$ if the positive threshold was reached, or append $(t^{*},\mathbf{u},-1)$ and set $S_{\mathrm{ref}}\leftarrow S_{\mathrm{ref}}-C^{-}_{\mathbf{u}}$ otherwise. The procedure repeats on the remaining interval $(t^{*},\,t_0+\Delta t]$, up to $N_{\max}$ events per pixel per step.

\subsection{Adaptive rendering}
\label{sec:accum_stats}
Crossing-time prediction is least reliable near disocclusions and mixed visibility. The compositing loop exposes three diagnostic signals at negligible extra cost. The transmittance rate $|\dot{\alpha}(\mathbf{u},t)|$ (rapid visibility change), the contributor count $c(\mathbf{u},t)$ (depth competition), and the coverage proxy $b(\mathbf{u},t)$ (accumulated blending mass). We combine them into a per-pixel risk score
\begin{equation}
r(\mathbf{u},t)=\mathrm{clip}\!\left(
w_b\,(1-\mathrm{norm}(b)) + w_c\,\mathrm{norm}(c) + w_{\alpha}\,\mathrm{norm}(|\dot{\alpha}|)
\right),
\label{eq:risk_score}
\end{equation}
and use $r$ to shrink the step size via $\Delta t \leftarrow \Delta t\,\kappa(r)$ with $\kappa(r)\in(0,1]$, and to suppress event emission when $b<b_{\min}$.

\subsection{Derivative-conditioned finite-bandwidth readout}
\label{sec:arbiter_short}

Real sensors serialize events through a finite-bandwidth readout, so localized bursts cause queuing delays, jitter, and drops.
We optionally model this with a tile-scanning arbiter: tiles maintain FIFO queues and a deterministic scanner releases at most $K_m$ events from tile $m$ per tick, with drop probability $p_{\mathrm{drop},m}$ and additive timestamp jitter $\sigma_{t,m}$.
We condition these on a tile-level activity score $\psi_m\in[0,1]$ aggregating the event-rate proxy $|\dot{L}|/C$, the transmittance rate $|\dot{\alpha}|$, and the contributor count $c$:
\begin{equation}
K_m = \left\lfloor K_0\,(1-\beta_K \psi_m)\right\rfloor,\qquad
p_{\mathrm{drop},m}=p_0+\eta\, \psi_m,\qquad
\sigma_{t,m}=\sigma_0+\gamma_\sigma\, \psi_m,
\end{equation}
where $K_0,p_0,\sigma_0$ are steady-state baselines and $\beta_K,\eta,\gamma_\sigma$ control the degradation rate. Under high activity, throughput falls while drops and jitter grow, reproducing realistic readout saturation.

\vspace{-4mm}

\section{Experiments} 
\label{sec:experiments}

\subsection{Experimental setting}
\label{sec:exp_setting}
We evaluate simulators along three axes: (i) event-stream fidelity against real events, (ii) event timestamp ratio metrics to analyze event bursts and (iii) synthetic-to-real transfer when training on simulated events and testing on real data. The results presented here are the average across all sequences in each dataset, while the supplementary material provides per-sequence results.

We compare TIDES against ESIM~\cite{Rebecq2018ESIM}, V2E~\cite{Hu2021v2e}, DVS-Voltmeter~\cite{Lin2022DVSVoltmeter}, and ICNS/IEBCS~\cite{IEBCS2023}.
We do not include PECS~\cite{Han2024PECS} as a baseline because its released pipeline targets a different scene-based setup than our shared-renderer protocol; evaluating PECS within our shared 4DGS-renderer protocol would require disabling its native physical image-formation pipeline, yielding a comparison that is methodologically misaligned and potentially unfair to PECS. Furthermore, PECS uses its own benchmarking protocol isolating other methods against itself.

We report a fixed-budget setting with a 10$\times$ query schedule for all methods and additionally ablate our adaptive rendering setting for TIDES as a separate row of results.
Our adaptive pose-sampler varied greatly in the frequency of sub-pose requests for each scene. Baseline methods struggled to operate with non-linear timestamp intervals at inference time, and performed universally worse when combined with our adaptive rendering.

\subsection{Datasets and shared renderer}
\label{sec:datasets_poses}
We evaluate on four real datasets spanning ego-motion and dynamic-scene regimes: EDS and DSEC capture ego-motion, while HS-ERGB and BS-ERGB contain dynamic objects~\cite{HidalgoCarrio2022EDS,Tulyakov2021TimeLens,Tulyakov2022TimeLensPP,Gehrig2021DSEC}.
When 6-DoF trajectories are provided, we use the official poses; otherwise, we estimate camera motion and sparse geometry from RGB using COLMAP for static scenes and Easi3R for dynamic scenes~\cite{Wang2024DUSt3R,Chen2025Easi3R}.
For each sequence we fit a dynamic Gaussian splatting scene from which we can render log-luminance $L(\mathbf{u},t)$ at any timestamp, providing a shared continuous-time rendering source for all simulators.
Frame-driven baselines are given renders at regular intervals, while step-based methods query only at their chosen times.
This isolates differences in time-stamping, thresholding, readout modeling, and occlusion handling from interpolation artifacts in the inputs.



\subsection{Metrics}
\label{sec:motion_scaled_metric}
We report three complementary metric families that target distinct simulator failure modes: (i) inter-event timing statistics, (ii) spatiotemporal alignment to real events, and (iii) batching/burst distortions.

Event generation can be characterized by the distribution of inter-spike intervals (ISI) $\tau$ at each pixel.
Following DVS-Voltmeter-style analysis, we fit an inverse-Gaussian (IG) model to observed ISIs and report the negative log-likelihood (IG-NLL) of real events under the simulator’s ISI distribution.
Lower IG-NLL indicates that the simulator reproduces the fine-grained timing of events rather than only their spatial support as a batch.

To measure whether simulated events occur at the correct location and time, we compare polarity-matched real and simulated event sets in a motion-scaled spatiotemporal embedding.
Each event $(x,y,t)$ is mapped to $\mathbf{e}=(x,y,vt)$, where $v$ (pixels/s) converts time errors into an expected spatial displacement so that a temporal offset $\Delta t$ is penalized as $v\Delta t$.
We compute a bidirectional nearest-neighbour (Chamfer-style) distance between the embedded point sets over fixed temporal windows, with voxel-grid density control to reduce sensitivity to repeated near-duplicate events.
Lower values indicate better spatiotemporal agreement without requiring explicit correspondences.

To diagnose timestamp artifacts that are visually evident as ``layered'' event bands, we report:
(i) Same-ts, the fraction of events sharing identical timestamps within a window (higher implies stronger batching);
(ii) ISI spike ratio, the fraction of ISIs near zero or concentrated at discrete clock quanta (indicating over-quantized timestamps);
and (iii) Fano-factor error, which measures whether the simulator matches real burst dispersion by comparing the Fano factor $F=\mathrm{Var}(N)/\mathbb{E}[N]$ of event counts in fixed spatiotemporal bins:
\begin{equation}
\mathcal{E}_{\mathrm{Fano}} = \frac{1}{|\mathcal{B}|} \sum_{b \in \mathcal{B}} \left|\log \frac{F_b^{\mathrm{pred}}}{F_b^{\mathrm{gt}}}\right|.
\end{equation}
Lower Same-ts, lower ISI spike ratio, and lower $\mathcal{E}_{\mathrm{Fano}}$ indicate reduced batching and more realistic burst statistics.
Full definitions and implementation details (binning, windowing, and normalization) are provided in the supplementary material.

\subsection{Downstream tasks}
To measure downstream task transfer, we train a range of models on simulated events and evaluate on real data. The specific tasks and models we use are: (1) E2VID~\cite{rebecq2019e2vid} for reconstruction, (2) TimeLens-XL~\cite{ma2024timelensxl} for video frame interpolation (VFI), (3) Depth Any Event Stream~\cite{depth_any_events} for depth estimation, and (4) ESS on DSEC~\cite{dsec_segmentation} for segmentation. For VFI, we use the image reconstructed from the ground-truth event stream as reference and report PSNR and LPIPS. Additional motion-task results (flow and odometry) are reported in the supplementary tables.

\begin{figure}[thp]
    \centering
    \includegraphics[width=\textwidth]{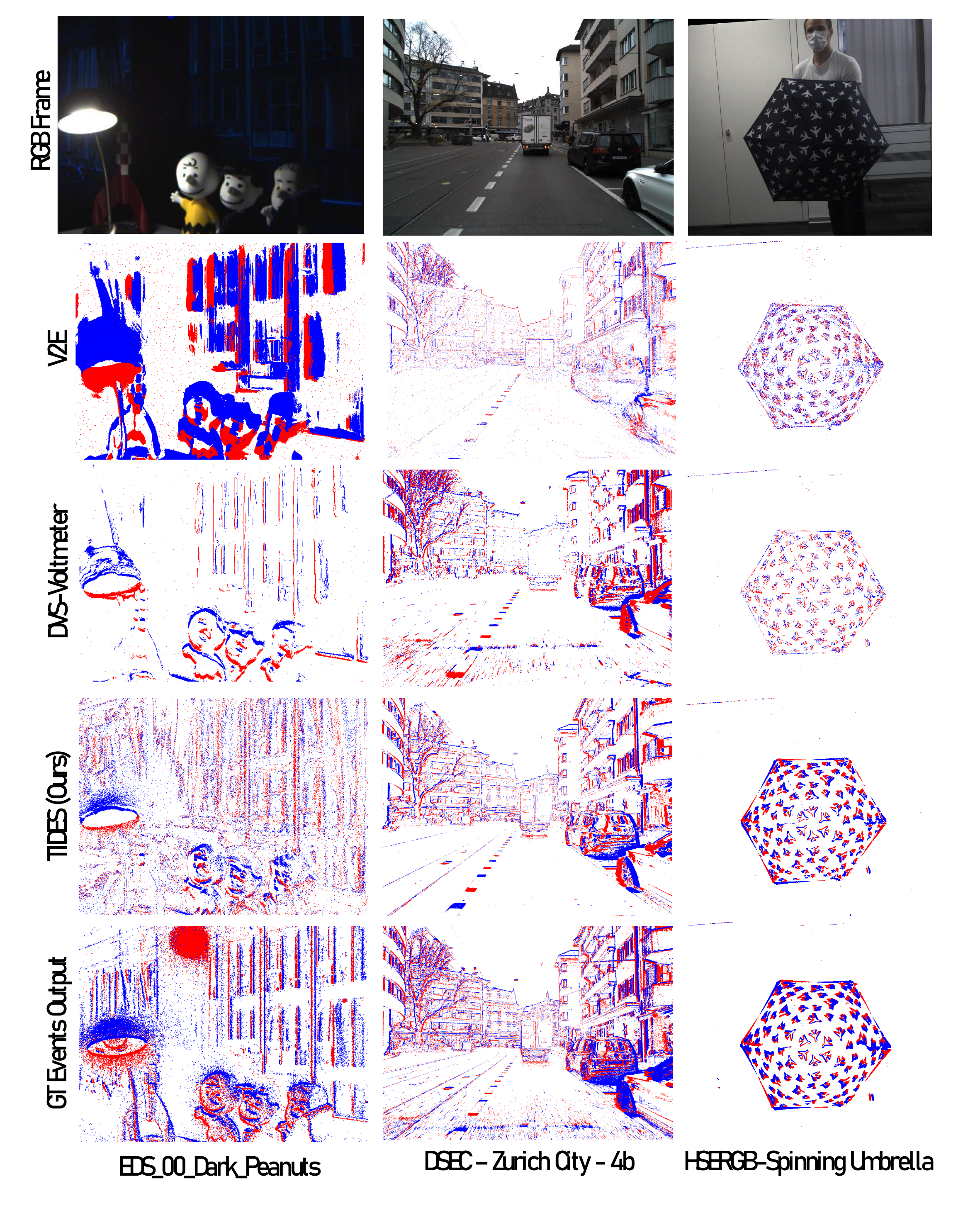}
    \caption{Illustrations of the event frames created from the ground truth, TIDES and popular event simulation methods V2E and ESIM}
    \label{event_images}
\end{figure}

\begin{figure}[thp]
    \centering
    \includegraphics[width=0.95\textwidth]{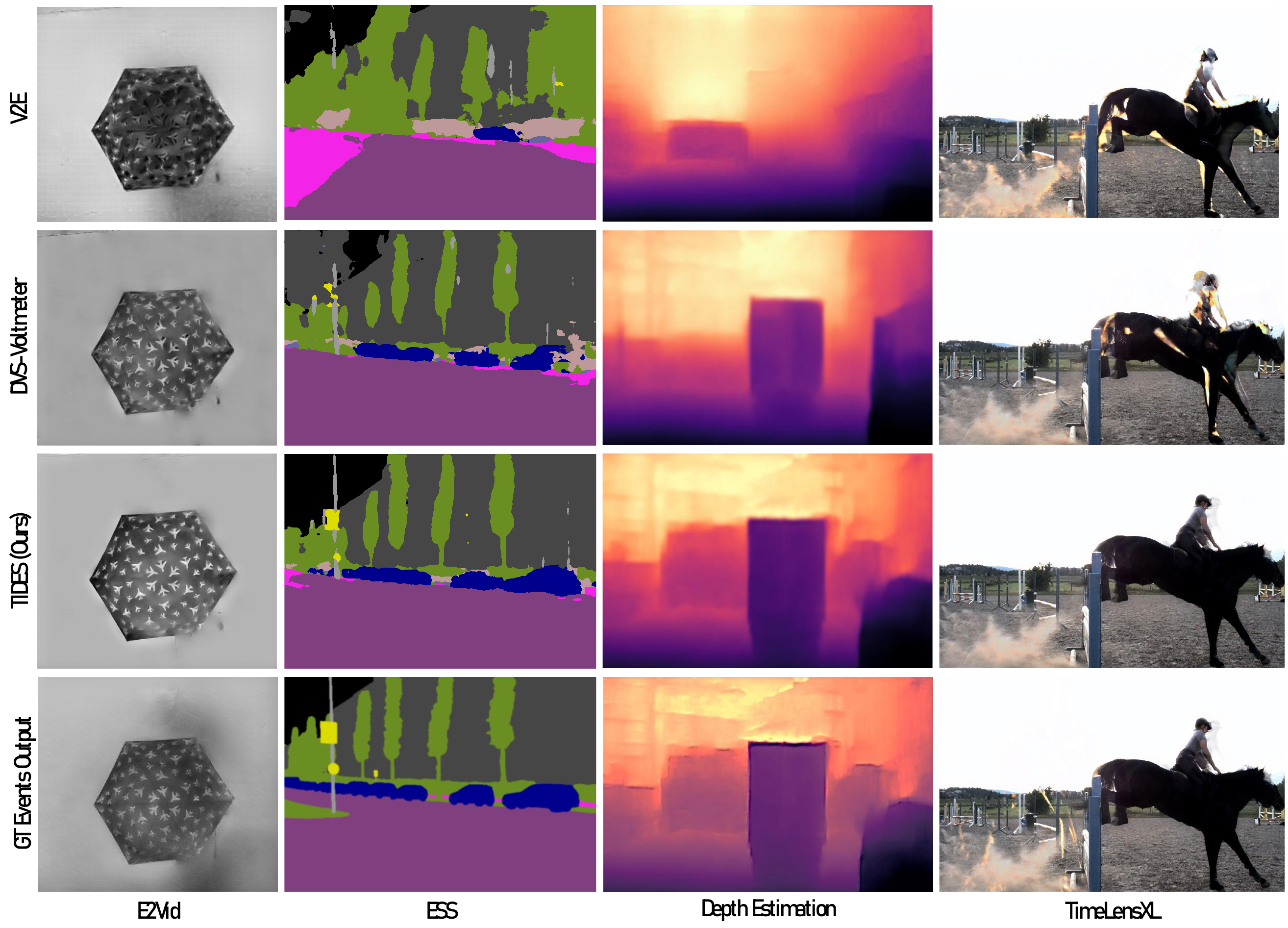}
    \caption{Visualizations of the downstream models using individual event generations including the ground truth event.}
    \label{exp1}
\end{figure}

\subsection{State-of-the-art comparison}
\label{sec:main_results}
Table~\ref{tab:fidelity_all} and Table~\ref{tab:supp_batching_metrics} report event-stream fidelity and timestamp-burst diagnostics, while Table~\ref{tab:downstream_summary} reports downstream transfer.

\begin{table*}[t]
\centering
\caption{Event fidelity per dataset. Best, second-best, and third-best per column are highlighted in red, orange, and yellow, respectively (excluding TIDES (adaptive) from color ranking).}
\small
\setlength{\tabcolsep}{4.5pt}
\resizebox{\textwidth}{!}{%
\begin{tabular}{lcc|cc|cc|cc}
\toprule
& \multicolumn{2}{c|}{\textbf{EDS}} 
& \multicolumn{2}{c|}{\textbf{HS-ERGB}} 
& \multicolumn{2}{c|}{\textbf{BS-ERGB}} 
& \multicolumn{2}{c}{\textbf{DSEC}} \\
\textbf{Simulator} 
& \textbf{IG-NLL} $\downarrow$ & \textbf{Chamfer} $\downarrow$
& \textbf{IG-NLL} $\downarrow$ & \textbf{Chamfer} $\downarrow$
& \textbf{IG-NLL} $\downarrow$ & \textbf{Chamfer} $\downarrow$
& \textbf{IG-NLL} $\downarrow$ & \textbf{Chamfer} $\downarrow$ \\
\midrule
ESIM                 & 0.004748 & 0.046932 & 0.006763 & 0.021335 & 0.007669 & 0.039233 & 0.096768 & 0.150414 \\
V2E                  & {\orangec 0.004058} & {\orangec 0.044124} & {\orangec 0.005823} & {\yellowc 0.016149} & {\yellowc 0.006092} & {\yellowc 0.035944} & 0.099792 & 0.153248 \\
ICNS/IEBCS            & {\yellowc 0.004575} & {\yellowc 0.046710} & 0.008341 & 0.028272 & 0.010608 & 0.045028 & {\yellowc 0.092576} & {\yellowc 0.142781} \\
DVS-Voltmeter         & 0.006050 & 0.050914 & {\yellowc 0.006388} & {\redc 0.014871} & {\orangec 0.005163} & {\orangec 0.033895} & {\orangec 0.080542} & {\orangec 0.133779} \\
\midrule
\textbf{TIDES (10x)}  & {\redc 0.003840} & {\redc 0.043245} & {\redc 0.004984} & {\orangec 0.014892} & {\redc 0.004392} & {\redc 0.033002} & {\redc 0.068641} & {\redc 0.121962} \\
\cdashline{1-9}
\textbf{TIDES (adaptive)}  & 0.003733 & 0.042622 & 0.002343 & 0.013772 & 0.004251 & 0.032260 & 0.068439 & 0.120611 \\
\bottomrule
\end{tabular}
}%
\label{tab:fidelity_all}
\vspace{-0.8em}
\end{table*}

\begin{table*}[t]
\centering
\caption{Batching diagnostics across datasets (average)}
\small
\setlength{\tabcolsep}{3.2pt}
\renewcommand{\arraystretch}{1.1}
\resizebox{\textwidth}{!}{%
\begin{tabular}{lccc|ccc|ccc|ccc}
\toprule
& \multicolumn{3}{c|}{\textbf{EDS}} & \multicolumn{3}{c|}{\textbf{HS-ERGB}} & \multicolumn{3}{c|}{\textbf{BS-ERGB}} & \multicolumn{3}{c}{\textbf{DSEC}} \\
\textbf{Simulator}
& \textbf{Same-ts} $\downarrow$ & \textbf{Fano} $\downarrow$ & \textbf{ISI spike} $\downarrow$
& \textbf{Same-ts} $\downarrow$ & \textbf{Fano} $\downarrow$ & \textbf{ISI spike} $\downarrow$
& \textbf{Same-ts} $\downarrow$ & \textbf{Fano} $\downarrow$ & \textbf{ISI spike} $\downarrow$
& \textbf{Same-ts} $\downarrow$ & \textbf{Fano} $\downarrow$ & \textbf{ISI spike} $\downarrow$ \\
\midrule
ESIM            & 0.281985 & {\yellowc 0.506379} & 0.096938 & 0.166213 & {\yellowc 3.305385} & 0.064902 & 0.279757 & {\yellowc 1.689136} & 0.111152 & 0.165757 & {\yellowc 3.296069} & {\yellowc 0.050237} \\
V2E             & {\yellowc 0.076272} & 4.284378 & {\orangec 0.047712} & 0.126701 & 6.409716 & 0.081733 & 0.155642 & 6.473020 & 0.089496 & 0.226925 & 6.495386 & 0.141100 \\
ICNS/IEBCS      & {\orangec 0.046909} & 2.417609 & {\redc 0.034290} & {\orangec 0.074536} & 4.975071 & {\orangec 0.049699} & {\orangec 0.102743} & 5.063936 & {\orangec 0.065669} & {\yellowc 0.114483} & 4.957434 & 0.073942 \\
DVS-Voltmeter   & 0.102028 & {\orangec 0.313285} & 0.061608 & {\yellowc 0.076948} & {\orangec 2.698272} & {\yellowc 0.051943} & {\yellowc 0.112803} & {\orangec 1.436576} & {\yellowc 0.069397} & {\orangec 0.060686} & {\orangec 3.155488} & {\orangec 0.039239} \\
\midrule
\textbf{TIDES (10x)}      & {\redc 0.040882} & {\redc 0.228440} & {\yellowc 0.052497} & {\redc 0.050215} & {\redc 2.069280} & {\redc 0.038454} & {\redc 0.085554} & {\redc 0.816516} & {\redc 0.041768} & {\redc 0.057469} & {\redc 2.332386} & {\redc 0.024181} \\
\cdashline{1-13}
\textbf{TIDES (adaptive)} & 0.025234 & 0.338062 & 0.055599 & 0.051124 & 2.017263 & 0.046340 & 0.080981 & 0.608790 & 0.033215 & 0.056436 & 2.199856 & 0.021470 \\
\bottomrule
\end{tabular}
}%
\label{tab:supp_batching_metrics}
\vspace{-0.9em}
\end{table*}

TIDES achieves the best fidelity across all four datasets under both a fixed 10$\times$ query budget and our adaptive budget (Table~\ref{tab:fidelity_all}). Gains are largest under fast motion and occlusions, where frame-driven simulators alias visibility changes and finite-difference slope estimates collapse events onto a few timestamps. In contrast, TIDES renders instantaneous, visibility-consistent $\dot{L}$ in the same compositing order as the primal image, enabling continuous-time threshold crossings (including multi-crossing per pixel) without choosing an arbitrary differencing interval.

Burst diagnostics in Table~\ref{tab:supp_batching_metrics} show that improvements come from corrected timestamp structure, not event-rate inflation like v2e typically highlights from its extreme batching. TIDES consistently reduces Same-ts batching and burst distortion (Fano, ISI spikes). Adaptive stepping further improves IG-NLL and Chamfer by refining time only in mixed-visibility regions indicated by $(\alpha,\dot{\alpha},b,c)$, with the largest gains on HS/BS-ERGB where pose sampling is sparse. Among baselines, V2E and ICNS/IEBCS improve realism via sensor/readout blocks, and DVS-Voltmeter models stochastic timing, but they remain limited by discretely sampled driving signals; TIDES removes this bottleneck by coupling continuous-time rendering with visibility-consistent derivatives and risk-aware time control. These results are exhasterbated by the visual event frames seen in Fig. \ref{event_images}.

TIDES produces smoother temporal dispersion along motion boundaries and fewer frame-aligned timestamp bands than frame-driven baselines, matching the reduced batching statistics in Table~\ref{tab:supp_batching_metrics}. 
These improvements translate to downstream transfer, models trained on TIDES events achieve the best overall performance across reconstruction, frame interpolation, depth, and segmentation (Table~\ref{tab:downstream_summary}), with qualitative examples shown in Fig.~\ref{exp1}.

\begin{table*}[t]
\centering
\caption{Quantitative downstream transfer (train on simulated, test on real). }
\small
\setlength{\tabcolsep}{5.0pt}
\renewcommand{\arraystretch}{1.15}
\resizebox{\textwidth}{!}{%
\begin{tabular}{l|cc|cc|cc|c}
\toprule
& \multicolumn{2}{c|}{\textbf{E2VID (HS-ERGB)}}
& \multicolumn{2}{c|}{\textbf{VFI (BS-ERGB)}}
& \multicolumn{2}{c|}{\textbf{EDA (DSEC)}}
& \multicolumn{1}{c}{\textbf{ESS (DSEC)}}  \\
\textbf{Simulator}
& \textbf{PSNR} $\uparrow$ & \textbf{LPIPS} $\downarrow$
& \textbf{PSNR} $\uparrow$ & \textbf{LPIPS} $\downarrow$
& \textbf{RMSE} $\downarrow$ & \textbf{RMSE log} $\downarrow$
& \textbf{mIoU} $\uparrow$ \\
\midrule
ESIM            & 12.2321 & 0.3283 & 28.5777 & 0.0580 & 12.6627 & 0.3163 & \orangec 17.2170  \\
V2E             & \yellowc 16.3587 & 0.2469 & 28.1585 & 0.0616 & \yellowc 12.6108 & \yellowc 0.3125 & 7.5678 \\
ICNS/IEBCS      & 15.2137 & \orangec 0.2097 & \orangec 31.0376 & \orangec 0.0422 & 19.6507 & 0.8791 & 16.3643  \\
DVS-Voltmeter   & \orangec 17.7328 &  \yellowc 0.2254 & \yellowc 28.6220 & \yellowc 0.0514 & \orangec 10.9048 & \orangec 0.2865 & \yellowc 16.9245  \\
\midrule
\textbf{TIDES (ours)} & \redc 18.1587 & \redc  0.1997 & \redc  34.3054 & \redc  0.0335 & \redc  10.8160 &  \redc  0.2795 & \redc 17.6898  \\
\bottomrule
\end{tabular}
}%

\label{tab:downstream_summary}
\vspace{-0.9em}
\end{table*}

\subsection{Ablations}
\label{sec:ablations}
We ablate each module while keeping the rendering source, sensor parameters, and evaluation protocol fixed.

Table~\ref{tab:ablate_fidelity_speed} shows that visibility-consistent time-tangent derivatives are the largest contributor as replacing the time-tangent slope with finite differences increases IG-NLL and Chamfer substantially.
Adaptive control is the next most important factor, confirming that refining time in mixed visibility improves timing beyond a fixed query schedule.
Disabling multi-crossing or mixed-visibility masking also degrades fidelity, while removing the readout model causes a smaller but consistent drop.

\begin{table}[thp]
\centering
\caption{Ablations isolating TIDES components on EDS scenes.}
\small
\setlength{\tabcolsep}{5.0pt}
\resizebox{\linewidth}{!}{%
\begin{tabular}{lcc}
\toprule
Variant & IG-NLL $\downarrow$ & Chamfer $\downarrow$ \\
\midrule
Full TIDES & 0.00373 & 0.04262 \\
\midrule
Finite-diff slope ($\dot L$ by FD) & 0.00521 & 0.04801 \\
No multi-crossing (1 event max / step) & 0.00439 & 0.04487 \\
Fixed step size (no controller) & 0.00486 & 0.04672 \\
No risk-gated refinement (no $(\alpha,\dot\alpha,b,c)$ shrink) & 0.00458 & 0.04593 \\
No mixed-visibility masking (emit even when $b<b_{\min}$) & 0.00422 & 0.04431 \\
No $\dot \alpha$ cue & 0.00461 & 0.04711 \\
No readout model (ideal timestamps) & 0.00388 & 0.04294 \\
\midrule
Compute-matched TIDES (adaptivity off, fixed \#queries) & 0.00405 & 0.04371 \\
\bottomrule
\end{tabular}
}%

\label{tab:ablate_fidelity_speed}
\vspace{-0.9em}
\end{table}

\vspace{-4mm}
\section{Conclusion}
We presented TIDES, a continuous-time event simulator built on dynamic Gaussian splatting that predicts per-pixel threshold crossings without discretizing time.
TIDES renders log-luminance and its instantaneous derivative in the same visibility ordering, making threshold-crossing prediction well-posed without the timestamp batching artifacts prevalent in rendering-based simulators.
Across four paired RGB-event benchmarks, TIDES achieves state-of-the-art fidelity and demonstrates stronger synthetic-to-real transfer on downstream event-vision tasks than existing frame-based simulators.
The quality of events simulated through TIDES is ultimately bounded by the fidelity of the underlying 4DGS scene reconstruction. However, our technique is agnostic to any particular 4DGS technique. Future efforts to improve the 4DGS rendering substrate will naturally translate to TIDES ongoing improvement.
\newpage
\newpage
\clearpage
\clearpage
\appendix
\newpage
\newpage
\clearpage
\clearpage
\bibliographystyle{splncs04}
\bibliography{main}
\end{document}